\documentclass{article}
\usepackage{ijcai23}

\usepackage{times}
\usepackage{soul}
\usepackage{url}
\usepackage[hidelinks]{hyperref}
\usepackage[utf8]{inputenc}
\usepackage[small]{caption}
\usepackage{graphicx}
\usepackage{amsmath}
\usepackage{amsthm}
\usepackage{booktabs}
\usepackage{algorithm}
\usepackage{algorithmic}
\usepackage[switch]{lineno}

\title{Learning and Reasoning Multifaceted and Longitudinal Data for Poverty Estimates and Livelihood Capabilities of Lagged Regions in Rural India}

\author{
Atharva Kulkarni$^1$\and
Raya Das$^2$\and Ravi S. Srivastava$^3$\and 
Tanmoy Chakraborty$^4$
\affiliations
$^1$Language Technologies Institute, Carnegie Mellon University, USA \\
$^2$Indian Council for Research on International Economic Relations, New Delhi, India \\
$^3$Centre for Employment Studies, Institute for Human Development, Delhi, India \\
$^4$Indian Institute of Technology Delhi, India\\
\emails
$^1$atharvak@cs.cmu.edu,
$^2$rdas@icrier.res.in,
$^3$ravi.srivastava@ihdindia.org,
$^4$tanchak@iitd.ac.in
}

\begin{document}

\maketitle

\begin{abstract}
    Poverty is a multifaceted phenomenon linked to the lack of capabilities of households to earn a sustainable livelihood, increasingly being assessed using multidimensional indicators. Its spatial pattern depends on social, economic, political, and regional variables. Artificial intelligence has shown immense scope in analyzing the complexities and nuances of poverty. The proposed project aims to examine the poverty situation of rural India for the period of 1990-2022 based on the quality of life and livelihood indicators. The districts will be classified into `advanced', `catching up', `falling behind', and `lagged' regions. The project proposes to integrate multiple data sources, including conventional national-level large sample household surveys, census surveys, and proxy variables like daytime, and nighttime data from satellite images, and communication networks, to name a few, to provide a comprehensive view of poverty at the district level. The project also intends to examine causation and longitudinal analysis to examine the reasons for poverty. Poverty and inequality could be widening in developing countries due to demographic and growth-agglomerating policies. Therefore, targeting the lagging regions and the vulnerable population is essential to eradicate poverty and improve the quality of life to achieve the goal of `zero poverty'. Thus, the study also focuses on the districts with a higher share of the marginal section of the population compared to the national average to trace the performance of development indicators and their association with poverty in these regions.

\end{abstract}

\section{Introduction}

Poverty is a complex situation in which the lack of capabilities translates to low income of the household \cite{nussbaum1993quality}. From a monetary perspective, poverty can be described as an interlacement of income distribution below a threshold value and the disproportions that exist within that boundary \cite{balaji2020negotiating}. It is a state of destitution in which individuals lack the basic essential means, such as food, water, shelter, and money, required to sustain their daily livelihood \cite{sharathchallenges}. While poverty has been a perennial socio-economic problem of mankind, the last two decades have witnessed a steady decline in global poverty \cite{world2020poverty}. However, owing to the COVID-$19$ pandemic, the compounding effects of climate change and socio-economic conflicts,  the pursuit to end poverty has suffered a significant setback for the first time in a generation. The pandemic resulted in the addition of almost $100$ million people diving into extreme poverty \cite{world2020poverty}. Moreover, as the United Nations (UN) lists poverty eradication as one of its primary Sustainable Development Goals (SDGs), global communities are striving hard to develop efficient techniques for poverty tracking, estimation, and eradication. 

\paragraph{\bf India under crisis:} As per Agriculture Census, 2015-2016\footnote{\url{https://agcensus.nic.in/document/agcen1516/T1_ac_2015_16.pdf}}, around 68\% of the population resides in rural India. The latest Multidimensional Poverty Index (MPI) scores indicate that the poverty score is $32.75\%$  among the rural population, contrary to $8.8\%$ in urban India. The BIMARU\footnote{BIMARU is an acronym formed from the first letters of the names of the Indian states of Bihar, Madhya Pradesh, Rajasthan, and Uttar Pradesh.} states continued to have the most deprived districts of the country, with some statistics being comparable to Sub-Saharan African countries, thus, calling into question the inclusiveness of policies in India. Around $50\%$ population of the country is engaged in agriculture and allied sector; therefore, the regional development of agriculture has an immense impact on the quality of life of rural households. Indian society has a hierarchy with different levels of mobility across social groups. As per the NFHS-4 data, two minority classes, namely scheduled caste (SC) and scheduled tribe (ST) households, have more poverty prevalence than general social groups. 

\paragraph{\bf Poverty estimation is crucial:} For any nation, accurately measuring poverty statistics and the economic characteristics of the population critically influences its research and national policies \cite{vskare2016poverty}. Moreover, economic growth cannot be the exclusive goal of a nation's economic policies; it is equally vital to ensure that the benefits of economic prosperity reach all segments of society. This underpins the importance of assessing poverty in all its manifestations. Also, poverty measurement is crucial to evaluate how an economy is performing in terms of providing a certain minimum standard of living to all its citizens. 
 In summary, measuring poverty has significant implications for policy drafting and implementation. 
As a result, it is no surprise that poverty estimation and tracking have garnered the attention of economists, social scientists, and statisticians alike.

\paragraph{\bf Multifaceted measurement of poverty:} Poverty, however, is not just based on the monetary distribution of wealth amongst the masses but is a multifaceted idea comprehensive of social, financial, and political components. Thus, to unthread the fabric of poverty and understand its nuances, a deeper and multidimensional study of its different facets is required. Such {\em multidimensional measurement of poverty} encompasses two approaches -- poverty as capability deprivation, and poverty as a measure of deprivation \cite{atkinson2003multidimensional}. The Multidimensional Poverty Index (MPI), jointly developed by the Oxford Poverty and Human Development Initiative (OPHI) and United Nations Development Programme, considers both these factors (incidence and intensity of deprivation) for measuring poverty. The widely adopted Alkire and Foster’s methodology \cite{alkire2011counting} considers the three indicators of standard of living, education, and income at the household levels to measure multidimensional poverty. However, in the case of developing countries, such as India, one must look beyond these aspects, as here, poverty estimation is beset by several quantitative and qualitative concerns. This calls for the consideration of other ancillary factors, such as the variance in climate, healthcare facilities, dietary habits, political status, cultural influence, infrastructure development, and geographical benefits and drawbacks along with the financial information \cite{ahluwalia1980benefits}.


\begin{figure}[!t]
    \centering
    \includegraphics[width=\columnwidth]{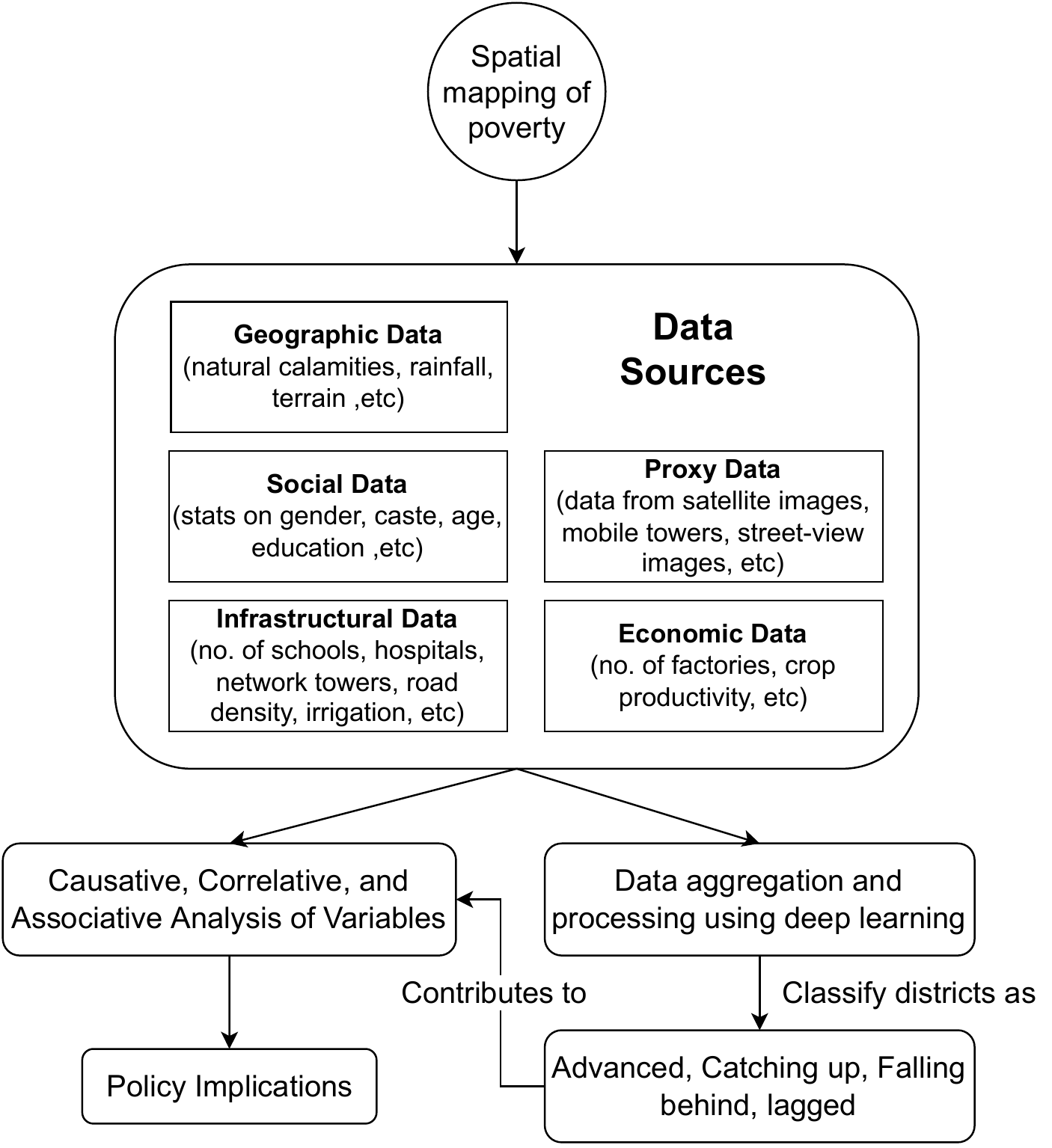}
    \caption{A bird's-eye view of our proposal.}
    \label{fig:overallmodel}
    \vspace{0mm}
\end{figure}

\paragraph{\bf Existing studies on India-specific poverty estimation and research gaps:} For India, the traditional narrative on poverty estimation makes use of publicly-available data collected through household surveys, such as the National Family and Health Survey (NFHS) ~\cite{mohanty2011multidimensional,chaudhuri2013multi,mishra2013multi}, National Sample Survey (NSS) \cite{sarkar2012multi,mishra2013multi}, Indian Human Development Survey (IHDS)  \cite{dehury2015regional} to name a few, to calculate the multidimensional poverty index. The census survey of the government of India publishes village-level statistics, and ICRISAT publishes the longitudinal village-level data on land relations, crop yield, access to facilities, and resource possession. However, due to the paucity and the lack of comparability, veracity, and precision of the data at different time periods, accurate estimation and tracking of poverty are challenging. Thus, recent studies in the realm of poverty estimation have focused on alternate and proxy data sources, such as satellite images \cite{jean2016combining,xie2016transfer,ayush2020generating,Ayush_Uzkent_Tanmay_Burke_Lobell_Ermon_2021}, mobile phone usage \cite{smith2014poverty,blumenstock2015predicting,smith2021estimating}, geospatial information \cite{puttanapong2020predicting}, and information on the web \cite{sheehan2019predicting}. Along with being inexpensive to produce and easier to scale, the use of these data sources also introduces the scope for utilizing modern machine learning (ML)  techniques for poverty and quality of life estimation. While the use of artificial intelligence (AI) is a regular feature for achieving other sustainable development goals, such as climate change, education, healthcare, clean energy, and gender equality, its application to poverty tracking has been limited. Moreover, studies on poverty estimation in India have exclusively focused on household survey data, excluding the possibility of using alternative proxy data sources. 

\paragraph{\bf Longitudinal and temporal analysis of poverty estimate:} While India has witnessed a commendable rise in its economic growth post-independence, it has been non-inclusive and exclusionary, exacerbating the rural-urban divide, regional disparities, and social and gender-based inequalities \cite{dev2008inclusive}. Studies on analyzing multidimensional poverty in India highlight the disparity at regional levels, where eastern and central states reflect high MPI scores \cite{das2021multidimensional}. The states of Jharkhand, Uttar Pradesh, Rajasthan, Odisha, Bihar, Chhattisgarh, and Arunachal Pradesh, have a higher poverty head-count ratio while Kerala, Mizoram, Nagaland, Punjab, Himachal Pradesh, and Haryana have lower poverty rate \cite{tripathi2020measurement}. Moreover, the temporal distribution of poverty has not been constant, as, over time, the poverty has aggrandized in the states of Manipur and Arunachal Pradesh in the rural setting, and Meghalaya, Odisha, and Jharkhand in the urban setting \cite{tripathi2020measurement}. This underlines the need for temporal and longitudinal analysis of poverty estimation. With the amalgamation of traditional as well as proxy data sources, such investigations will help unravel the peculiarities of areas diving south of the poverty line, thus, enabling the policymakers to establish suitable and appropriate poverty eradication guidelines. Similarly, on other hand, identifying the regions where the poverty index has increased over time will help diagnose the policies that worked and the ones which did not. Lastly, it will also help differentiate the geographical regions of chronic poverty from transient poverty.

\paragraph{\textbf{Major contributions:}} 
The proposed study investigates the spatial mapping of poverty by the quality of life and livelihood capabilities in the lagged regions vs. advanced regions of rural India at the district level. We also intend to a {\em longitudinal and temporal analysis} to focus on the trajectory of growth and catching up capabilities of lagged regions of the country. We further herald a new research direction towards estimating poverty in lagged Indian states using data aggregation and integration and expound on its application using ML. Figure \ref{fig:overallmodel} shows a bird's eye view of our overall proposal. 
Thus, with the poverty-struck Indian states as a case study, we aim to address the following broad research questions: 
\begin{enumerate}
\setlength{\itemsep}{0 pt}
    \item {\bf Multidimensional data integration:} How different types of proxy data sources be used to estimate poverty in India?
    \item {\bf Efficient prediction:} How can these proxy data sources be integrated with the traditional survey data for more accurate, interpretable, and efficient poverty estimation through traditional ML and advanced neural models?
    \item {\bf Longitudinal analysis and estimation:} Can a temporal and longitudinal examination of poverty using proxy data sources reveal salient information about the factors that contribute to poverty trajectory?
    \item {\bf Correlation, association, and  causality analysis:} Can we identify variables that have a direct causal, correlative, and associative effect on poverty estimation? 
    \item {\bf Policy Implications:} How can causal, correlative, and associative analysis, along with the classification and forecasting study help in policymaking? 
\end{enumerate}

\section{Goals}
The primary purpose of this research is to demonstrate the shortcomings of the conventional approach to estimating poverty in India and to illustrate new approaches and methodologies for doing so with the recent advances in AI and ML. Furthermore, this study emphasises the need for a temporal and longitudinal analysis of poverty, rather than a static and time-constrained one, in order to fully comprehend the expression and expansion of poverty in the socio-economically backward states of India. This section expands on these goals.

\subsection{Limitations of Traditional Data Sources}
The data from the household expenditure and income surveys, such as the National Family and Health Survey (NFHS), different rounds of the National Sample Survey (NSS), and the Indian Human Development Survey (IHDS), form the backbone for identifying and measuring the poverty status of Indian households. Although the census data provides a comprehensive measurement of the material living standard of individuals in a population, it is conducted over long cycles (usually every ten years) and encompasses only a few characteristics of the target population. Household surveys, on the other hand, cover a wide range of variables, but their reliability for local statistical inference is limited. Conducting such surveys is also expensive, time-consuming, imprecise, and at times infeasible for poverty assessment. The majority of these statistics are generated over a reasonably long period and do not accurately reflect the attributes that affect the livelihoods of the residents of a particular area. Moreover, the target population is also not the same for all the datasets. For example, the states of Arunachal Pradesh, Bihar, Jharkhand, and Orrisa have a considerable tribal population along with the standard rural and urban distribution. This raises concerns over the credibility of the estimated poverty figures. With a few exceptions, such as information about education and job status, such surveys focus solely on household monetary information \cite{alkire2011counting}. However, as poverty is not a uni-dimensional phenomenon and not just a mere reflection of the economic status of an individual, it encompasses other concomitant aspects, such as data on infrastructural development, agricultural growth, vehicle count, network towers, political climate, caste information, electricity usage, and so forth. Such elements are more relevant in rural settings as they are direct markers of prosperity. While household data comprising the income statistics does have its place in poverty prediction, they have certain limitations, as stated above. Therefore,  {\em we propose a data integration mechanism that takes into account both the traditional datasets along with the new proxy data sources for poverty prediction}.

\subsection{Towards the Use of Proxy Data Sources}
There have been numerous studies on multidimensional poverty estimation, albeit most of them incorporate just the conditions of education, health, and living standards in addition to the economic dimension \cite{alkire2011counting}. While these strategies are unquestionably superior to the single-source poverty assessment methodologies, they do have some detriments, as discussed in the previous section. To compensate for these deficiencies, the advent of big data, combined with technological breakthroughs in ML, offers great promise for poverty tracking as well as interpreting and predicting social-economic conditions. This accounts for the data gathered from social media, remote sensing, agricultural growth, vehicular traffic, infrastructural development, mobile phone meta-data, and housing details. Political, cultural, and environmental aspects also factor in while estimating poverty. There are examples aplenty that elaborate on the successful use of these proxy data sources for poverty estimation across geographies. For instance, remotely sensed images, such as Landsat data and night-time light images, serve as the most representative data sources as they provide important information about the region's landscape. Various case studies on China \cite{niu2020measuring,shi2020identifying}, Thailand \cite{puttanapong2020predicting}, Philippines \cite{hofer2020applying}, Bangladesh \cite{steele2017mapping}, and African countries \cite{jean2016combining,Ayush_Uzkent_Tanmay_Burke_Lobell_Ermon_2021} have demonstrated the usability of remote sensing for poverty tracking and estimation. Some other studies have shown that employing data from mobile phones \cite{blumenstock2015predicting}, communication networks \cite{smith2014poverty}, and political climate \cite{van2006trends} also yields promising results. While these research works do validate the use of proxy data sources for poverty estimation, they also bring to light that no such study has scrutinized the Indian subcontinent. Moreover, these works use independent proxy data sources for poverty estimation and do not contemplate the idea of combining and integrating different data sources for more efficient, explainable, and robust predictions. Thus, this works aims to address this bottleneck by proposing a data aggregation and integration methodology, combining the traditional as well as proxy data sources, for poverty estimation of Indian states. In addition, we offer a multi-input deep learning-based architecture for aggregating and processing data from many sources. Finally, we provide methods for investigating the correlative, associative, and causative relationships between various input variables and poverty estimation, the identification of which can help in better policymaking.

\subsection{Temporal Analysis of Poverty}
Poverty is a temporal phenomenon that aggravates or declines over time. Thus, to forecast poverty statistics and its progress in a region, its historical data characterizing the temporal shifts should also be taken into consideration. While such a methodology would, in all likelihood, yield better, more accurate, and robust predictions, it will also help differentiate the prominent factors contributing towards poverty from the weaker ones. It will also help diagnose the schemes, plans, and policies that succeeded during a specific time leading to poverty alleviation or that had an adverse effect, leading to poverty expansion. The contemporary literature on poverty estimation exclusively focuses on static and time-specific data, failing to account for the temporal aspect of poverty. In rapidly developing countries like India, it is paramount to take into account the temporal dimension of poverty to pinpoint its core causes and design policies to tackle it. Thus, this research proposes a temporal data collection, integration, and prediction scheme for more robust poverty forecasting.


\section{Methodology}

In this section, we elaborate on the different proxy data sources, the data integration methodology, the use of ML and deep learning techniques for poverty estimation, and the temporal analysis of poverty. Furthermore, we classify the districts based on the performance (MPI score) between 1993-2015 considering several rounds of NFHS household surveys into `advanced', `catching up', `falling behind', and `lagged' districts. The study further targets the four most lagged states -- Uttar Pradesh, Jharkhand, Bihar, and Odisha, as they rank the highest in the multidimensional poverty index of 2021  \cite{tripathi2020measurement}.

\subsection{Proxy Data Sources and Data Integration}
To develop a functional methodology that governments of developing countries can use for accurate poverty estimation and tracking, one requires a dataset that is representative of the country’s population, that can be collected and timely updated automatically, and that is available at a fine level geographical granularity. In this section, we explore the different proxy data sources for poverty estimation as well as elaborate on the data integration approach.

\subsubsection{Remote sensing}
Remote surveying with satellite imagery is a low-cost and dependable method of tracking human development at fine spatial and temporal resolution \cite{jean2016combining}. Remote sensing involves various types of satellite imaging, conveying different types of information during the day and night. Daytime satellite images provide a wealth of information on the region's geography, infrastructure development, and population growth. Moreover, it can be used to infer other signals of prosperity, such as growth in the road network, building density, forest cover, and infrastructural expansion. The nighttime luminosity information provided by the satellite images provides a lens over a region's nocturnal activity. It serves as an ideal proxy for electricity consumption,  degree of electrification and population growth \cite{ghosh2013using}. Studies also showed a positive correlation of nighttime luminosity with carbon dioxide emissions, GDP, GDP per capita, constant price GDP, non-agricultural GDP, and capital stocks \cite{addison2015nighttime}. Therefore, we aim to collect and utilize both daytime and nighttime satellite image data as each has its own advantages. We propose to use the satellite imagery from the Landsat 7 mission from the years 2001, 2011, 2016, and 2019 to track daytime activity. The nighttime light data can be procured from the  United States Air Force Defense Meteorological Satellite Program (DMSP). The satellite's Operational Linescan System (OLS) sensors have a spatial resolution that allows them to make observations ranging from entire continents to less than a square kilometre. To map villages and districts to their satellite-image locations, we also propose to collect the information on each of their \textit{shapefiles}. Once a region has been linked to its satellite images, we can extract its human development attributes from the satellite-image features for that region, as stated above, based on its daytime and nighttime visual look from space.

\subsubsection{Communication networks}
Active research in the last decade has shown that information from mobile phone usage and telecommunications networks are a strong indication of a region's socio-economic status \cite{soto2011prediction,smith2014poverty,pokhriyal2015virtual}. We can automatically infer proxy indicators of poverty from unobtrusively obtained call network data by a detailed examination of patterns inherent in mobile phone users' collective behaviour. For instance, cellphone top-up behaviour suggests that poorer people are likely to top-up their phone credit regularly in small amounts, whereas wealthier people are more likely to top-up infrequently in larger amounts. Furthermore, an increase in network density of communication reflects an infrastructural development, an increase in population, and socio-economic well-being. In our study, {\em we propose to collect the Call Detail Records} (CDR) {\em data aggregated to the cell tower level}. The mobile phone operators collect such user-specific data primarily for billing purposes. Using such data, we can obtain a wealth of information about each call or text message, including the time, duration, caller and callee IDs, as well as the base station towers routing the call or text. Such precise data not only reveal the degree of the penetration rate of mobile technology in developing countries but also provides a relatively unbiased picture in terms of demographics. To protect users' privacy, we plan to collect the CDR data aggregated by the cell towers through which the calls are routed rather than using the data at an individual user level. {\em We plan to extract two types of data from the CDR, the first pertaining to a single tower} ({\em measuring the number of incoming/outgoing calls}), {\em and the second concerning a pair of two towers} ({\em measure the flow of calls between them}). The raw CDR data contains each cell tower's location information in latitude and longitudes. We intend to work at the spatial granularity of the {\em Voronoi areas} associated with cell tower placements. To successfully use such data, telecommunication providers just need to share anonymized, aggregated call detail records in a regulated manner. Early hints of this are already visible (e.g., the D4D Challenge\footnote{\url{http://www.d4d.orange.com/home}}), and different frameworks are being developed to attract even more providers to join the endeavour \cite{parate2009framework}.

\subsubsection{Other data sources as indicators of poverty}
Data from remote sensing and communication networks form the primary proxy data source for poverty estimation. Other variables that may contribute towards poverty are the environmental, political, and cultural characteristics of a region. For instance, natural calamities, such as floods, droughts, tornadoes, extreme/high low rainfall, and famines could result in poverty aggravation. Furthermore, the details about the ruling party of a region, the policies it has implemented, and quantifying its progress could also shed light on the region's economic development. Population-related statistics, such as the number of women, men, children, and senior citizens in an area might as well have some bearing on its economic condition. Information from local news snippets could also shed light on regional developments, crime rates, and other advancements. It also remains an open question whether cultural information, such as characteristics of different tribes, the different caste and their distribution, plays an active part in poverty tracking, assessment, and expansion. Street-view images of an area over time could also help depict its developments or degradation. {\em We intend to consider these ancillary factors and determine whether any of these positively correlate with poverty deprivation}.

\subsubsection{Data integration}
The distinctive aspect of this study is that, rather than relying solely on new proxy data sources for poverty estimation, we integrate them with traditional census and household survey data. The survey datasets provide vital individual-level information about age, sex, religion, caste, mother tongue, marital status, education, disability status, land ownership, irrigation infrastructure, and tenancy status, to name a few. They also include details on the availability of bathrooms, drinking water, separate kitchen, electricity, electronic devices, vehicle count, cooking fuel, and roof, wall, and floor materials, among other things\footnote{\url{https://censusindia.gov.in/census_and_you/data_item_collected_in_census.aspx}}. This results in a diverse dataset containing images, text, numeric data, and categorical labels. While traditional statistical methods cannot accommodate non-numeric data inputs, the current state-of-the-art deep learning architectures are perfectly suited for processing and merging unstructured data from disparate sources. We elaborate on this in the following section.

\subsection{Learning and Reasoning  Poverty Estimation}

\begin{figure*}
    \centering
    \includegraphics[width=13cm]{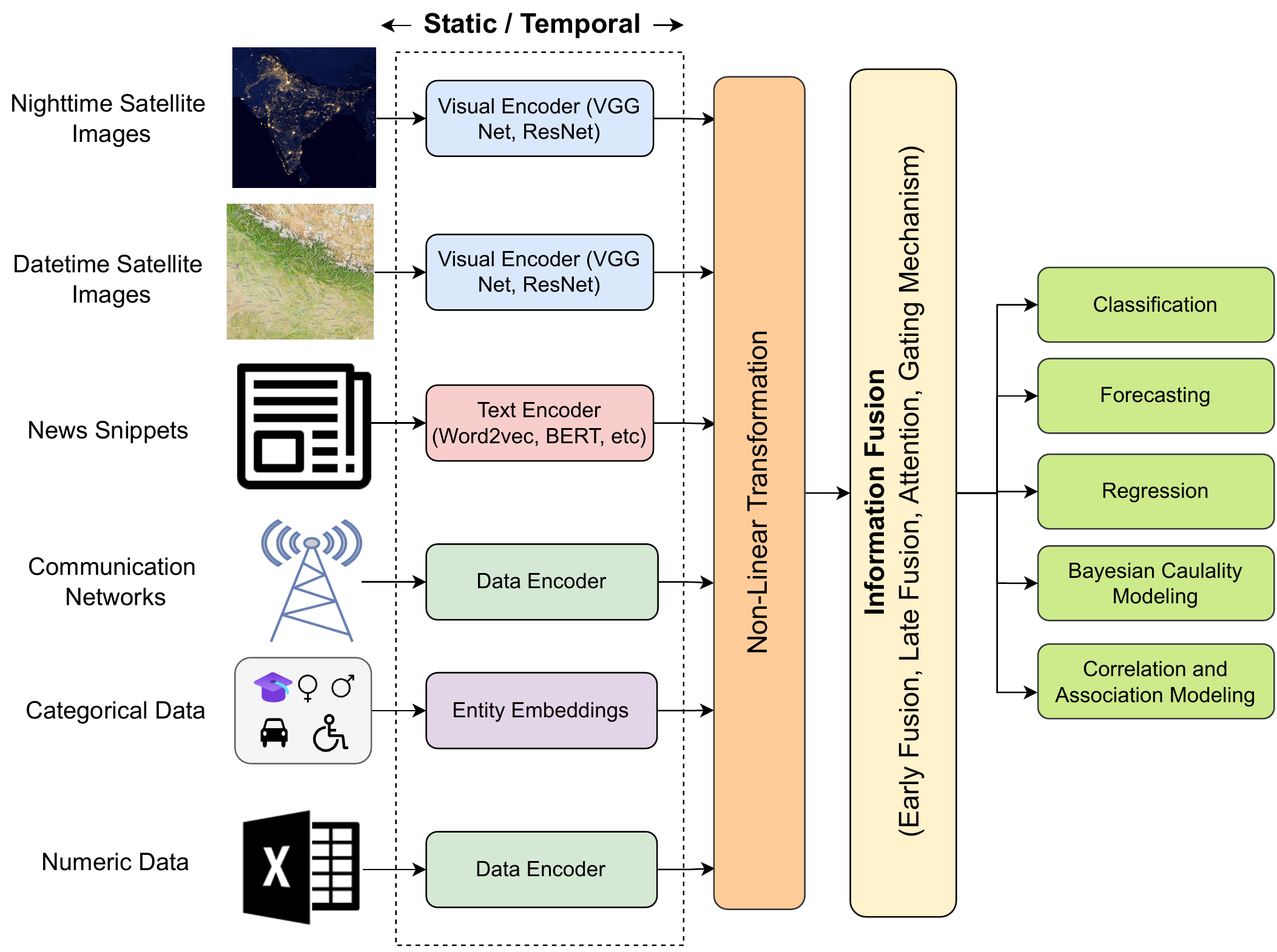}
    \caption{Proposed multi-input deep learning model for aggregating and processing data from proxy and traditional data sources.}
    \label{model}
\end{figure*}


\subsubsection{Neural models for poverty estimation}
Deep neural models are being aggressively employed to attain various sustainable development goals \cite{vinuesa2020role,vinuesa2021interpretable}. For poverty assessment and tracking as well, there have been numerous successful attempts of using ML methods utilizing traditional
\cite{isnin2020does,niu2020measuring} as well as proxy data sources \cite{jean2016combining,smith2014poverty}. However, none of these methods designs systems that aggregate and combine different data sources. To address this, {\em we aim to present a multi-input neural architecture that can aggregate and combine information from several disparate data sources in a non-trivial fashion}. Figure~\ref{model} presents a schematic diagram of our generalized deep neural model that takes various inputs, such as images, text, as well as numeric and categorical data. The  numeric data derived from the census and household surveys can be easily fed to a deep neural model by combining all the features into a vector representation. Categorical attributes, such as gender, education, religion, caste, and marital status can be encoded using a technique, called {\em entity embeddings} \cite{guo2016entity}. Unstructured data types of images and text snippets cannot be fed directly into neural models. While deep learning models specialize in processing unstructured data, they require input in the form of numeric representations. However, as we have limited training data, training visual or textual embeddings from scratch will, in all likelihood, not yield desirable results. Thus, {\em we propose to use a transfer learning approach wherein we use pre-trained models trained on large corpora and fine-tune them to our task-specific  dataset}. For the visual inputs of satellite images, we can utilize pre-trained CNN-based models, such as VGG Net \cite{simonyan2014very}, or ResNet \cite{He_2016_CVPR}. For the textual inputs, we prefer the language models, such as BERT \cite{devlin-etal-2019-bert} or RoBERTa \cite{liu2019roberta}. Each of the representations goes through a non-linear transformation before the data merging operation. The data merging can be done in various ways, such as early fusion, late fusion, naive concatenation, attention mechanism \cite{bahdanau2014neural}, or gating-based fusion. We treat our problem as a classification task wherein we have four ground-truth labels at the district level -- `advanced', `catching up', `falling behind', and `lagged'. This is calculated based on aggregating the individual MDPI scores at the district level.

\subsubsection{Correlation, Association, and Causality}
Analysing the relationship of the input variables with the target output can be a challenging step but it is important for strategic actions. Such insights are important to determine the driving factors (causative), factors that exhibit linear relationships (correlative), and factors that co-occur (associative). We propose to use the Bayesian Networks for identifying causative factors, Pearson's correlation to determine the correlative factors, and the Hypergeometric test to discover the associative factors. Such correlative, associative, and causal analyses will aid in shaping and designing efficient policymaking.

\subsection{Temporal Analysis for Poverty Estimation}
While the current vogue is to predict poverty using static data properties, this approach does not encapsulate the concept of poverty in its entirety. Poverty is a complex phenomenon with a very strong temporal component. Furthermore, recent studies have demonstrated that ML models trained for poverty prediction have poor time transferability, i.e., whether models built on data from one year can make sound predictions on data from another year \cite{bansal2020temporal}. Thus, for efficient and robust poverty assessment, we must consider and process data in a temporal fashion. For example, a consistent decline in rainfall in an area over time might actively contribute toward increased chronic poverty. However, outlier events, such as natural calamities or socio-political upheavals are rare phenomena leading to only ephemeral impoverishment and transient poverty as people finally recover. Thus, in this project, {\em we propose a new research direction and methodology for the temporal assessment and prediction of poverty using longitudinal dissection of data}. To start with, we plan to consider a time frame of $5$ years for the temporal analysis. State-of-the-art sequences learning models, such as LSTMs and GRUs have been very successful in capturing temporal dependencies and sequential data. The range of transformer-based architectures \cite{vaswani2017attention} succeed likewise in this using the self-attention mechanism. Using these architectures, we will train a neural network using {\em teacher-forcing} in which, during training, the model receives the ground-truth output $y(t)$ as input at time $t + 1$. Thus, for each time step, the model receives the standard inputs, along with the output from the previous time step. 

\section{Model Evaluation}

When evaluating a deep learning model for multifaceted and longitudinal data for poverty estimates and livelihood capabilities, it is important to consider several criteria. For our case we focus on the following six criteria:
\begin{enumerate}
    \item {\bf Accuracy:} The model should be able to accurately predict poverty estimates and livelihood capabilities based on the given data.
    \item {\bf Bias:} The model should not be biased towards any particular group or demographic. The model should be fair and impartial in its predictions.
    \item {\bf Generalizability:} The model should be able to generalize well to new data that it has not seen before. This is important because poverty estimates and livelihood capabilities can vary across different regions and populations.  Thus, the model needs to be able to capture these variations.
    \item {\bf Interpretability:} The model should be interpretable, meaning that it should be possible to understand how the model arrived at its predictions. This is important for ensuring that the model is not making predictions based on irrelevant or biased factors. Moreover, to design effective policies, the model interpretability can help provide causal relations between the input data features and the poverty prediction outcome.
    \item {\bf Robustness:} The model should be robust to changes in the data and any noise or errors that may be present. This is important because poverty estimates and livelihood capabilities can be affected by various factors, and the model should be able to handle these variations.
    \item {\bf Scalability:} The model should be scalable, meaning that it should be able to handle large volumes of data efficiently. This is important because poverty estimates and livelihood capabilities data can be vast and complex, and the model should be able to handle this complexity.
\end{enumerate}


\section{Challenges}
Though poverty amelioration has been a primary priority of the Indian government, the fact of the matter remains that millions of Indians continue to be poor by national and international standards, despite persistent efforts since independence. The utilization of AI/ML techniques may introduce additional challenges for efforts regarding poverty assessment. They are elaborated as follows.
\begin{enumerate}
    \item{ \bf Lack of labeled data:} Deep learning models require large amounts of labeled data to achieve accurate predictions. The more data you feed them, the better results they are likely to yield. However, in many cases, poverty-related data is not readily available or is difficult to label accurately. Therefore, this requires substantial efforts in data gathering, storing, and processing.
    \item{ \bf Data quality:} Poverty-related data can be subject to quality issues such as missing values, errors, and inconsistencies, leading to biased results. or example, there may be missing data on income or consumption, or the data may not be representative of the population being studied.
    \item{ \bf Data bias:} Data bias can occur when the training data used to train deep learning models is not representative of the population being studied. This can lead to inaccurate predictions and exacerbate existing inequalities.
    \item{ \bf Data privacy:} Poverty-related data may contain sensitive information about individuals or households, making it difficult to collect and share. This can limit the availability of data for training deep learning models.
    \item{ \bf High Carbon Footprint:} the AI algorithms processing big data have high energy requirements and carbon footprints, which can have a detrimental impact on SDG 7 (Affordable and Clean Energy) and SDG 13 (Climate Action). 
    \item{ \bf Lack of skilled personnel:} Finally, designing and monitoring AI-based systems necessitates experts in these fields. Moreover, as AI/ML is a rapidly evolving discipline, the recruited employees must regularly update their understanding of this field. However, India, which otherwise produces thousands of IT workers each year, has a severe scarcity of AI skills professionals.
\end{enumerate}


\section{Risks, Limitations, Ethical Considerations}
Poverty is a structural phenomenon, and tracing its causal factors over a time period might not be enough to articulate the causes of deprivation. The macro statistics, including budget allocation by government, public and private investment, and revenue generation of respective states, are not included in the analysis. Qualitative information like life history, challenges, and opportunities at the household level are not incorporated into the analysis. The data integration might reduce the quality and explainability of the individual datasets. The data used in the study are accessible from public platforms. For satellite images, outputs will be published at the district level. 


\section{Expected Results and Long-Term Plans}
The holistic results of this study will provide a cogent understanding of the underlying causes of poverty and the capability of lagged regions to come out of it. Socio-economic factors in terms of access to amenities, education, health, connectivity, and living conditions facilitate livelihood capabilities in rural India. Inability to generate sustainable livelihood results in pushing the lagged region behind. The proxy indicators have the potential for a deeper understanding of poverty. The traditional poverty measurement indicators of income level or consumption expenditure also have the probability of reporting error, whereas the analysis of poverty from different facets will assist policymakers in eradicating poverty. Even the advanced states have intrastate differences, and the districts with a concentration of marginalized populations are the most lagged regions of the state. The integrated results can provide a different perspective of poverty based on the quality of life and livelihood capabilities and strengthen the outcome of conventional indicators. Policy suggestions will be based on the causal relationship of socioeconomic determinants of quality of life based on longitudinal data. In this context, the current social welfare and rural development schemes will be evaluated based on the results. 


\section{Conclusion}
Regional poverty has been a major concern since the Independence period in India. The polarised growth in advanced states does not have the ripple effect of growth for low-income states. The proposal aims to understand the social, economic, regional, and cultural dimensions of uneven development and its impact on the poverty condition of the lagged region. The application of AI in targeting lagged regions has immense scope to identify the explaining factors of poverty. The analyses also trace the quality of life and development indicators of the advanced region and the catching-up region, which will suggest a path of development for the lagged region. Data integration is a powerful method to measure poverty rather than using only traditional or proxy variables. Poverty and inequality would be widening in developing countries due to an increase in population and growth-centred policies. Therefore, targeting the lagged region and the vulnerable population is essential to eradicate poverty and improve the quality of life to achieve the goal of ``zero poverty''. 

\clearpage
\newpage
{
\bibliographystyle{named}
\bibliography{ijcai23}}

\end{document}